\documentclass[10pt,twocolumn,letterpaper]{article}

\usepackage[pagenumbers]{iccv} 

\definecolor{lightyellow}{RGB}{255,255,250}
\definecolor{darkyellow}{RGB}{255,255,170} 
\definecolor{darkblue}{RGB}{0,0,255}     
\definecolor{darkpurple}{RGB}{200,0,200}  
\definecolor{darkred}{RGB}{230,0,0}        
\definecolor{darkgreen}{RGB}{0,128,0}     
\usepackage{tcolorbox}  
\usepackage{pythonhighlight}

\definecolor{iccvblue}{rgb}{0.21,0.49,0.74}
\usepackage[pagebackref,breaklinks,colorlinks,allcolors=iccvblue]{hyperref}

\def\paperID{8888888} 
\def\confName{ICCV}
\def\confYear{2025}

\title{
Multimodal Generation of Animatable 3D Human Models with AvatarForge}

\author{Xinhang Liu\\
HKUST
\and
Yu-Wing Tai\\
Dartmouth College
\and
Chi-Keung Tang\\
HKUST
}

\begin{document}
\maketitle
\begin{abstract}
We introduce \emph{AvatarForge}, a framework for generating animatable 3D human avatars from text or image inputs using AI-driven procedural generation. While diffusion-based methods have made strides in general 3D object generation, they struggle with high-quality, customizable human avatars due to the complexity and diversity of human body shapes, poses, exacerbated by the scarcity of high-quality data. Additionally, animating these avatars remains a significant challenge for existing methods. AvatarForge overcomes these limitations by combining LLM-based commonsense reasoning with off-the-shelf 3D human generators, enabling fine-grained control over body and facial details. 
Unlike diffusion models which often rely on pre-trained datasets lacking precise control over individual human features,
AvatarForge offers a more flexible approach, bringing humans into the iterative design and modeling loop,
with its auto-verification system allowing for continuous refinement of the generated avatars, and thus promoting high accuracy and customization. Our evaluations show that AvatarForge outperforms state-of-the-art methods in both text- and image-to-avatar generation, making it a versatile tool for artistic creation and animation.
\end{abstract}    
\begin{figure}[t]
    \centering
    \includegraphics[width=\linewidth]{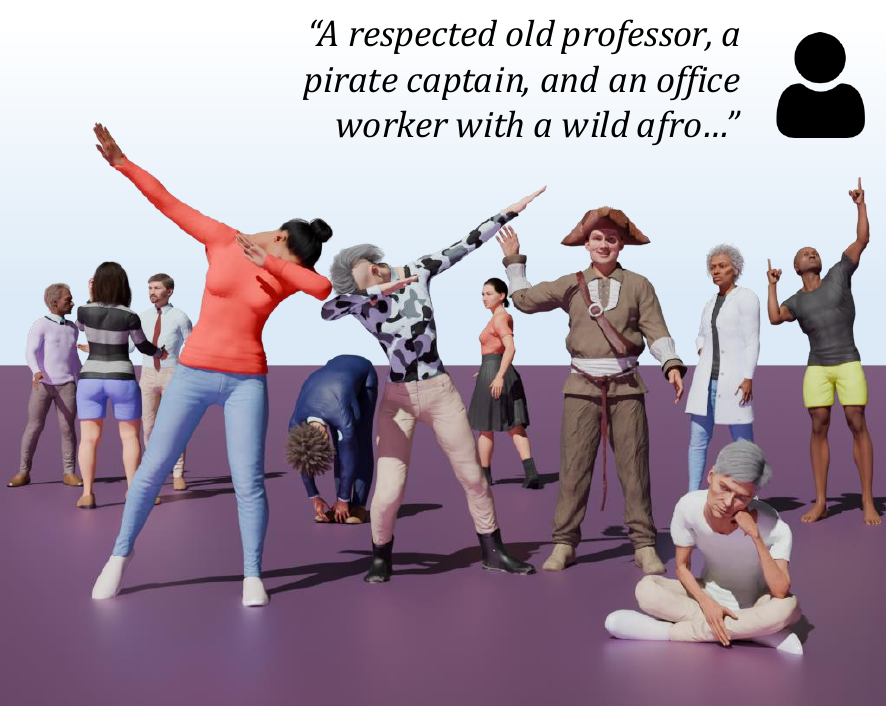}
    \vspace{-0.25in}
    \caption{\textbf{AvatarForge} for generating customizable and animatable 3D human avatars. The approach takes text or image inputs to create lifelike human figures with diverse body shapes, poses, and facial expressions. AvatarForge enables intuitive human modeling by refining the avatars according to user-specific requirements or feedback provided in natural language.}
    \vspace{-0.2in}
    \label{fig:teaser}
\end{figure}

\section{Introduction}
\label{sec:intro}
Despite substantial progress in general-purpose generative AI, both in 2D~\cite{ddpm, song2020denoising, rombach2022high, zhang2023adding} and 3D~\cite{poole2022dreamfusion, lin2023magic3d, hong2023lrm, zhang2024clay, siddiqui2024meta}, creating realistic, animatable human avatars remains a significant challenge. Current 3D human generation models often fail to capture intricate facial features, accurate body shapes, and realistic clothing textures. The difficulty intensifies when attempting to animate these avatars while maintaining visual fidelity. These limitations hinder the broader adoption of such technologies in industries like film, gaming, and mixed reality, where demand for high-quality, dynamic human modeling is rapidly growing.

To address these challenges, we introduce \emph{AvatarForge}, a novel AI-driven framework for generating lifelike, animatable 3D human avatars from text and image inputs. At the core of \emph{AvatarForge} is an LLM agent that interacts with users to understand detailed specifications for body, facial features, and clothing. By leveraging off-the-shelf 3D human generators, the system mimics the nuanced workflows of experienced artists, making 3D avatar creation more accessible and efficient, even for non-experts.

While current large language models (LLMs)~\cite{achiam2023gpt, dubey2024llama, team2023gemini} excel at a wide range of tasks, they often struggle with the complexities of manipulating sophisticated 3D human parametric generators. To address this problem, we introduce a novel auto-verification agent that evaluates the generated 3D models against the input text and images. This agent ensures that the produced avatars align closely with user specifications, enhancing both the quality and accuracy of the results. The auto-verification process works by assessing the congruence between the generated model and the desired attributes (e.g., facial features, body shape, clothing). When discrepancies are detected, the system refines the model iteratively to improve its alignment with the input. Thus, this dynamic refinement process gives \emph{AvatarForge} a significant edge over conventional diffusion-based methods, which typically lack this level of fine-grained control and feedback. Additionally, \emph{AvatarForge} allows for the creation of dynamic, articulated 3D avatars that can assume various poses and motions. By integrating another LLM agent focused on human motion control, users can animate these models using natural language commands, further enhancing the versatility of the generated avatars.

Our experimental results demonstrate that \emph{AvatarForge} surpasses existing methods in both visual quality and customization. In summary, our contributions are:
\begin{itemize}
    \item We present \emph{AvatarForge}, a novel framework integrating LLM agents with 3D human generators for creating highly customizable, animatable human avatars.
    \item We enhance a dynamic manual procedure supported by agent auto-verification to help LLMs better operate 3D human generators.
    \item Our system enables the creation of dynamic, animatable 3D human models, with motion control via natural language.
    \item Experimental results show that \emph{AvatarForge} outperforms current state-of-the-art methods in visual quality.
\end{itemize}

\section{Related Work}
\label{sec:related}

\noindent\textbf{Text-to-3D Generation.} Text-to-3D generation has evolved with approaches ranging from direct 3D diffusion models to methods that generate 3D representations from multi-view 2D images. Early works~\cite{nichol2022point, jun2023shap, xu2023dmv3d} focused on training 3D diffusion models on datasets of captioned 3D assets. However, these methods are often limited by the availability and diversity of high-quality 3D data. To overcome these limitations, many methods have shifted to utilizing 2D priors from captioned images, with works such as DreamFusion~\cite{poole2022dreamfusion} optimizing Neural Radiance Fields (NeRF) to match renders to the belief of a pre-trained text-to-image model. Recent advances~\cite{lin2023magic3d, qian2023magic123, tang2023dreamgaussian,liu2024deceptivenerfy} have expanded on this idea, employing alternative 3D representations such as hash grids, meshes, and 3D Gaussians (3DGS). Furthermore, two-stage methods~\cite{liu2023one, long2024wonder3d, liu2023syncdreamer} use a text-to-image or -video model to first generate multiple views of a 3D object, followed by fitting these views using techniques like NeRF or 3DGS. While these methods demonstrate impressive capabilities, they are often constrained by the need for precise multi-view consistency and the limited control over the final 3D shape and texture of the generated models.

AvatarForge differs significantly by integrating LLM-driven procedural generation into the avatar creation pipeline. Rather than relying solely on pre-trained diffusion models, AvatarForge introduces a novel framework where commonsense reasoning via LLMs informs the avatar generation process, enabling fine-grained control over both the human body and facial features. This approach allows for greater flexibility and precision in generating 3D avatars from text or image prompts, offering more control over specific anatomical details and poses, and addressing the scalability challenges seen in prior methods.

\vspace{2mm}
\noindent\textbf{Text-to-3D Human Models.} For 3D human model generation, a variety of approaches have been explored, with many drawing from DreamFusion-style pipelines. Early works~\cite{kolotouros2023dreamhuman, jiang2023avatarcraft, NEURIPS2023_0e769ec2} employ the SDS loss for optimizing 3D human representations based on text-to-image priors. DreamHuman~\cite{kolotouros2023dreamhuman} and DreamWaltz~\cite{NEURIPS2023_0e769ec2} introduce pose-conditioning and skeleton-based supervision to refine the generated human bodies. However, these methods are often limited by issues such as the Janus effect (duplicating object parts) and content drift, along with the slow optimization process, which can take hours per model. Additionally, while DreamHuman leverages deformable NeRFs for human-specific modeling, it still suffers from limitations in handling detailed body and facial features, and often fails to address the diversity of human body shapes with the level of control AvatarForge offers.

In contrast, AvatarForge not only incorporates pose-conditioning through its LLM-based commonsense reasoning but also integrates an auto-verification system that enables continuous refinement of the generated avatars. This iterative feedback loop significantly reduces artifacts such as the Janus effect, ensuring high-quality, highly-customizable 3D human avatars. Furthermore, AvatarForge’s system operates in real-time with more efficient optimization, in stark contrast to the hours-long refinement typically required by other methods. AvatarForge also tackles the challenge of diverse body shapes and poses, addressing the scarcity of high-quality data more effectively by combining off-the-shelf 3D human generators with LLM-driven procedural generation, allowing it to generate highly varied avatars from both text and image inputs with greater accuracy and realism.

\vspace{2mm}
\noindent\textbf{LLM Agents and Visual Programming.} Recent developments in large language models (LLMs) have made significant strides in multimodal tasks, including text-to-image generation~\cite{alayrac2022flamingo, li2023blip} and complex reasoning tasks~\cite{brown2020language, ouyang2022training, team2023gemini}. LLMs have demonstrated impressive capabilities in zero-shot and few-shot learning, enabling them to process and generate complex outputs from textual and visual inputs. The integration of external tools like visual foundation models and computational frameworks further expands the abilities of LLMs, allowing them to tackle sophisticated tasks such as image generation, editing, and visual reasoning~\cite{schick2023toolformer, wang2024internvid}. However, despite these advancements, LLMs have not yet fully realized their potential in the domain of 3D human modeling, where high fidelity, accuracy, and control are paramount. Moreover, while LLMs are increasingly adept at handling multimodal tasks, they still face challenges when it comes to generating complex 3D structures, such as human avatars, where the intricacy of geometry, texture, and animation needs to be seamlessly combined.

AvatarForge pushes the boundaries of what is possible by combining LLM-driven commonsense reasoning with real-time procedural 3D generation, specifically for human avatars. While existing LLM-based frameworks are primarily focused on visual reasoning tasks, AvatarForge leverages LLMs to iteratively guide the avatar creation process and ensure high customization through user-driven refinement. This results in a much more interactive, user-centric approach to avatar generation compared to prior methods, which typically rely on pre-optimized models that are difficult to modify or personalize.

\section{Method}
\label{sec:method}
This section details the inner working mechanism of AvatarForge, focusing on how the LLM extracts semantic information from text and images, improves the output through chain-of-thought reasoning, evaluates the generated avatars via auto-verification, and finally animates these avatars using motion agents. Additionally, we present how we dynamically construct a manual generation system that adapts to the LLM's evolving capabilities and ensures real-time iterative refinement.

\subsection{Teaching LLMs Procedural Generation}

\noindent\textbf{Issues in LLM-Procedural Generator Integration.~} The integration of LLMs with procedural generators to create 3D human avatars introduces several challenges that we must overcome to enable effective avatar generation. Existing state-of-the-art procedural generators we use are equipped with an extensive array of adjustable parameters, each influencing the resulting avatar's features, from body shape to attire. This versatility, however, introduces high complexity, presenting a  problem for general-purpose LLMs, which are typically not trained to handle such specialized tasks. In summary the key challenges include:

\begin{itemize} \item \textbf{High Dimensionality:} Procedural generators involve a multitude of parameters and settings that can be overwhelming for LLMs. LLMs are generally optimized for processing textual or lower-dimensional data, making it difficult for them to navigate the high-dimensional space of procedural generation. \item \textbf{Specific Domain Knowledge:} Procedural generators require detailed knowledge of human anatomy, clothing, and facial features. LLMs, while powerful, do not inherently possess this specialized knowledge. Hence, they must be customized or fine-tuned to effectively use procedural generation tools for human avatars. \item \textbf{Feedback Loops:} The lack of immediate, explicit feedback in 3D avatar generation complicates the learning process for LLMs. Without real-time guidance or structured feedback, LLMs struggle to make the necessary adjustments in subsequent iterations to refine the generated avatars. \end{itemize}

To address these challenges, we introduce a dynamic feedback mechanism and iterative refinement system, as detailed in the following sections, which enable LLMs to overcome these issues and generate high-quality avatars.

\begin{figure}[t]
    \centering
    \includegraphics[width=\linewidth]{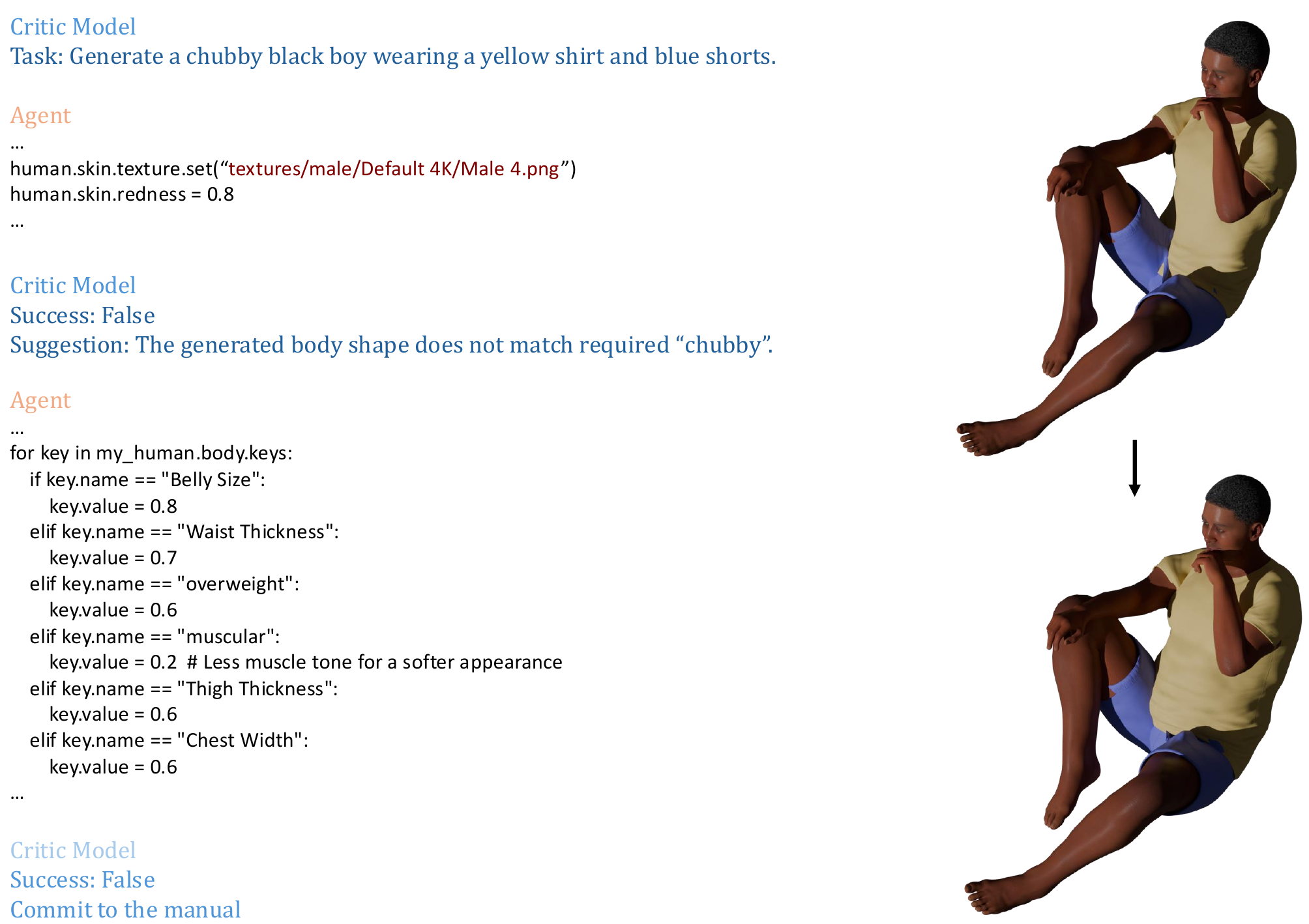}
    \caption{Iterative refinement process in the agent-critic framework for 3D human avatar generation. The LLM agent interacts with the critic model during the procedural generation of a human avatar. In this example, the agent's first attempts fail to meet the required criteria. The feedback loop and dynamic manual updates are key to refining the avatar generation process.}
    \vspace{-0.15in}
    \label{fig:manual}
\end{figure}

\begin{figure*}[t]
    \centering
   
{\footnotesize \begin{tcolorbox}[colframe=black, colback=lightyellow, coltitle=black, left=2mm, right=2mm]

\textcolor{darkblue}{\textbf{User}:} A basketball player.

\textcolor{darkpurple}{\textbf{Agent}:} 

\textbf{Observation:}
The user has requested the generation of a basketball player model. The description specifics like age, ethnicity, or specific physical attributes are not provided.

\textbf{Plan:}
Select a preset that generally matches the athletic build typical for a basketball player, assuming a male model unless otherwise specified. Given the lack of specifics about ethnicity, random choice among presets suitable for sports figures.
Adjustments will emphasize an athletic build, with particular attention to muscle tone and height; typical basketball player features include long limbs and a tall stature. For the facial features, a confident and focused expression will be attempted.
Default sports or casual clothing, possibly along with sports shoes, will be chosen.


\textbf{Code:}
\begin{python}
# Customizing body shape for athleticism
for key in my_human.body.keys:
    if key.name in ("muscular", "Shoulder Width", "Chest Muscles", "Biceps", "Traps Muscles", "Quad Muscles"):
        key.value = 0.8   # Highly defined muscles
    elif key.name in ("overweight"):
        key.value = 0.2   # Lean physique
    elif key.name in ("Thigh Thickness", "Calf Muscles"):
        key.value = 0.7   # Strong legs
    elif key.name in ("Height"):
        key.value = 200   # Setting height for a basketball player (around 200cm)

# Customizing facial features for a focused look
for key in my_human.face.keys:
    if key.name == "eye_tilt":
        key.value = -0.1   # Slight focus
    elif key.name == "browridge_loc_vertical":
        key.value = 0.2
    elif key.name == "jaw_width":
        key.value = 0.4    # Strong jawline
    elif key.name == "cheek_fullness":
        key.value = -0.2   # Leaner face

# Set height and age
my_human.height.set(value_cm=200)
my_human.age.set(25, realtime=False)

# Assign sporty clothing and footwear
my_human.clothing.outfit.set("outfits/male/Sports/Gym_Outfit.blend")
my_human.clothing.footwear.set("footwear/male/Sports/HG_Running_Shoe_2.blend")
\end{python}

\end{tcolorbox} }
\vspace{-0.1in}
 \caption{\textbf{Chain-of-Thought reasoning process for generating a basketball player avatar.} The figure illustrates the LLM's sequential steps: the observation of the input description, formulation of a plan for avatar creation, self-reminders to avoid potential bugs, and the implementation of the plan through Python code adjustments for customization. This structured process enables the LLM to generate an accurate and detailed avatar based on the user's input.}
 \vspace{-0.2in}
    \label{fig:cot}
    \end{figure*}

\vspace{2mm}
\noindent\textbf{Dynamic Manual Generation and Iterative Refinement.~} To help LLMs manage these complexities, we introduce a strategy of dynamic manual generation. This system creates a constantly evolving manual that serves as a guide for the LLM agent as it interacts with procedural human generators. Unlike static manuals, which are fixed and require frequent human intervention to update, the dynamic manual adapts in real time based on the LLM's interaction with the avatar generation process.

The manual is continuously updated by feedback from the auto-verification agent, which assesses the accuracy of the avatars generated by the LLM. Each time an avatar is generated, the verification agent evaluates it based on a set of criteria derived from the input (text or image), and if discrepancies are found, the manual is updated with guidance on how to avoid similar errors in the future. This process allows the LLM to progressively improve its ability to use the procedural generator, increasing the accuracy and diversity of the generated avatars as illustrated in Figure~\ref{fig:manual}.

\vspace{2mm}
\noindent\textbf{Discussion: Dynamic vs. Static Manual.~} The dynamic manual offers key advantages over traditional static manuals. Static manuals are rigid and require manual updates, limiting their ability to adapt to new challenges during the generation process. In contrast, the dynamic manual evolves with real-time feedback, ensuring the LLM always has the latest strategies and insights to handle unforeseen scenarios. This continuous refinement leads to more accurate guidance, allowing for more efficient avatar generation. Unlike static manuals, which need periodic updates and retraining, the dynamic manual’s adaptability enables ongoing improvement with minimal intervention.

\subsection{Extracting Semantic Information: LLM and Input Interpretation} AvatarForge’s ability to generate accurate 3D human avatars begins with the LLM's ability to interpret both textual and image-based inputs. For text-based inputs, the LLM parses the description and extracts key semantic attributes, such as body type, pose, clothing, and facial features. The LLM’s understanding of human anatomy and context allows it to transform this information into a detailed internal representation of the avatar.

When given image-based inputs, AvatarForge employs a vision-language model (VLM) to interpret the visual information. The input image is analyzed by a vision encoder, which converts the visual features into high-dimensional embeddings. These embeddings are then processed by the LLM to align them with semantic descriptors, such as body proportions and facial features. This dual input system enables AvatarForge to handle both textual and visual inputs, ensuring flexibility and versatility in the generation process.

\vspace{2mm}
\noindent\textbf{Chain-of-Thought and Incremental Refinement.~} The key to improving the generation process lies in chain-of-thought (CoT) reasoning, a technique where the LLM breaks down the task into logical, sequential steps. Rather than attempting to generate the avatar in one go, the LLM reasons through the process by asking, ``What features should the avatar have?" and ``How should it be posed?" This reasoning process enables the LLM to ensure that the avatar meets the desired criteria.

For example, upon receiving a prompt, the LLM might first identify key features (e.g., ``The character has short brown hair''), and then refine the avatar's pose (e.g., ``The character should be smiling and looking slightly to the right''). By engaging in this chain-of-thought reasoning, the LLM can assess the intermediate results and correct any mistakes before completing the avatar generation. Figure~\ref{fig:cot} shows an example of our CoT reasoning.

This process is especially important for handling edge cases, where the description may be ambiguous or the input includes conflicting attributes. The ability to reason incrementally and address issues as they arise leads to more accurate and natural outputs.

\subsection{Auto-Verification and Feedback Loops} The auto-verification agent plays a critical role in ensuring that the generated avatars align closely with the input criteria. Once the LLM generates an avatar, it is evaluated by the verification agent, which compares the 3D model to the original input (text or image). The verification is done based on several factors, such as anatomical accuracy, pose alignment, and fidelity to the input description.

The verification agent is designed to measure semantic similarity, assessing how closely the generated avatar matches the key attributes in the input. For example, if the input describes a ``tall man with a beard wearing a red jacket,'' the agent will check if the avatar's body shape, facial features, and clothing match the description.

If the avatar fails to meet the criteria, the verification agent provides detailed feedback that guides the LLM in refining the model. This feedback is used to update the dynamic manual, allowing the system to continuously learn and adapt to the task at hand. The iterative nature of this process results in improved avatar generation over time, as the system builds on its previous successes and corrects earlier mistakes.

\subsection{Animating the Generated Avatar} 

After generating the 3D human avatar, we animate it using the Motion-Agent framework~\cite{wu2024motionagent}. This conversational system enables the generation, editing, and understanding of human motion sequences. By integrating Motion-Agent, the framework allows for dynamic, multi-turn conversations to generate and refine complex motion sequences. Through simple textual instructions like “the character performs a backflip” or “slow down the last part of the movement,” users can guide the avatar's animation. The system interprets these commands and adjusts the motion accordingly, offering real-time customization and fine-tuning. This conversational approach enables highly interactive and detailed motion generation, making it easier to animate avatars with realistic and responsive movement in virtual environments.

\begin{figure*}[t]
    \centering
    \includegraphics[width=\linewidth]{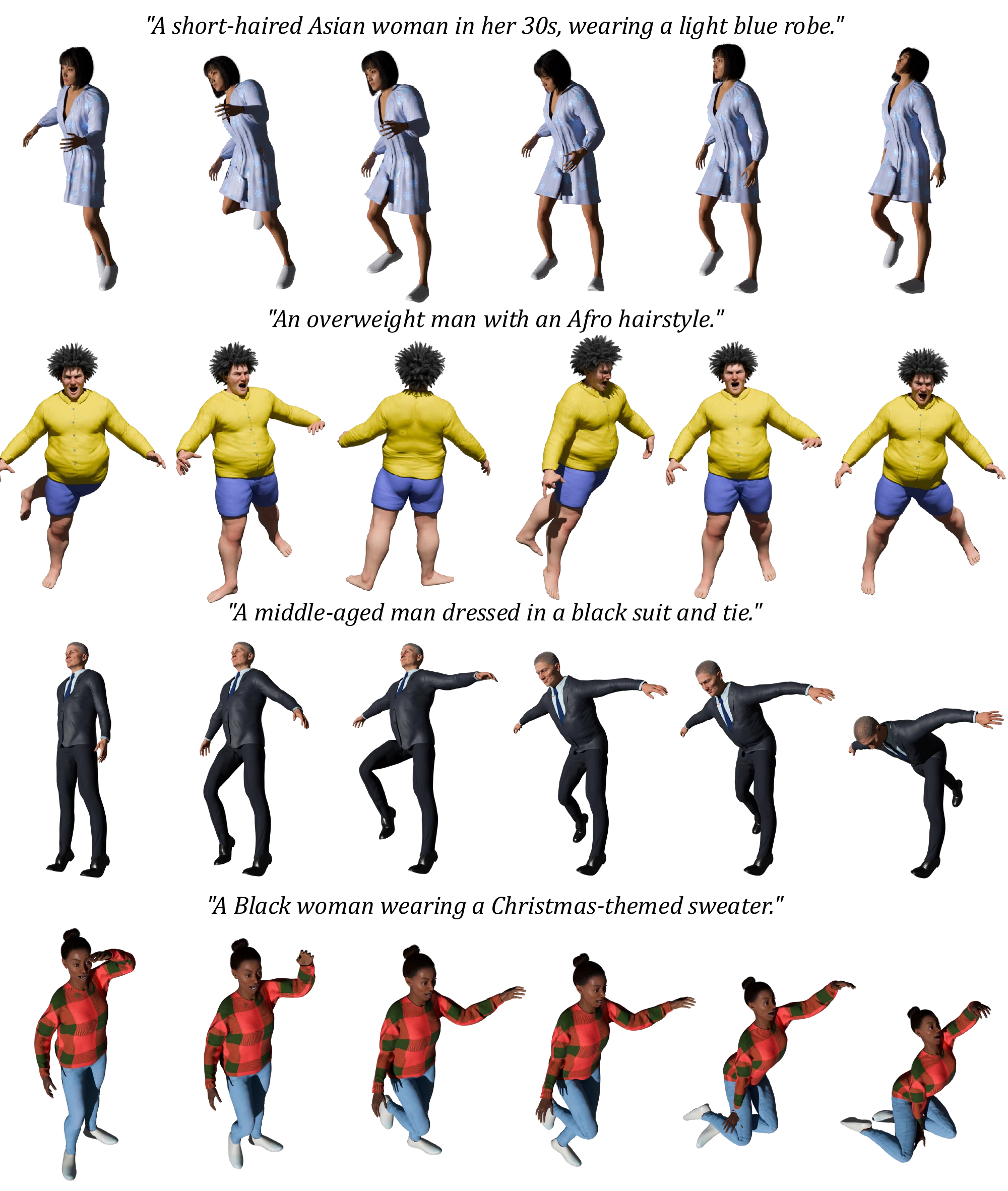}
    \caption{Diverse 3D human avatars generated using AvatarForge, showcasing a variety of body types, outfits, and poses based on text descriptions. This demonstrates the framework's capability to create highly customizable and realistic human models.}
    \label{fig:gallery}
\end{figure*}

\begin{figure}[t]
    \centering
    \includegraphics[width=\linewidth]{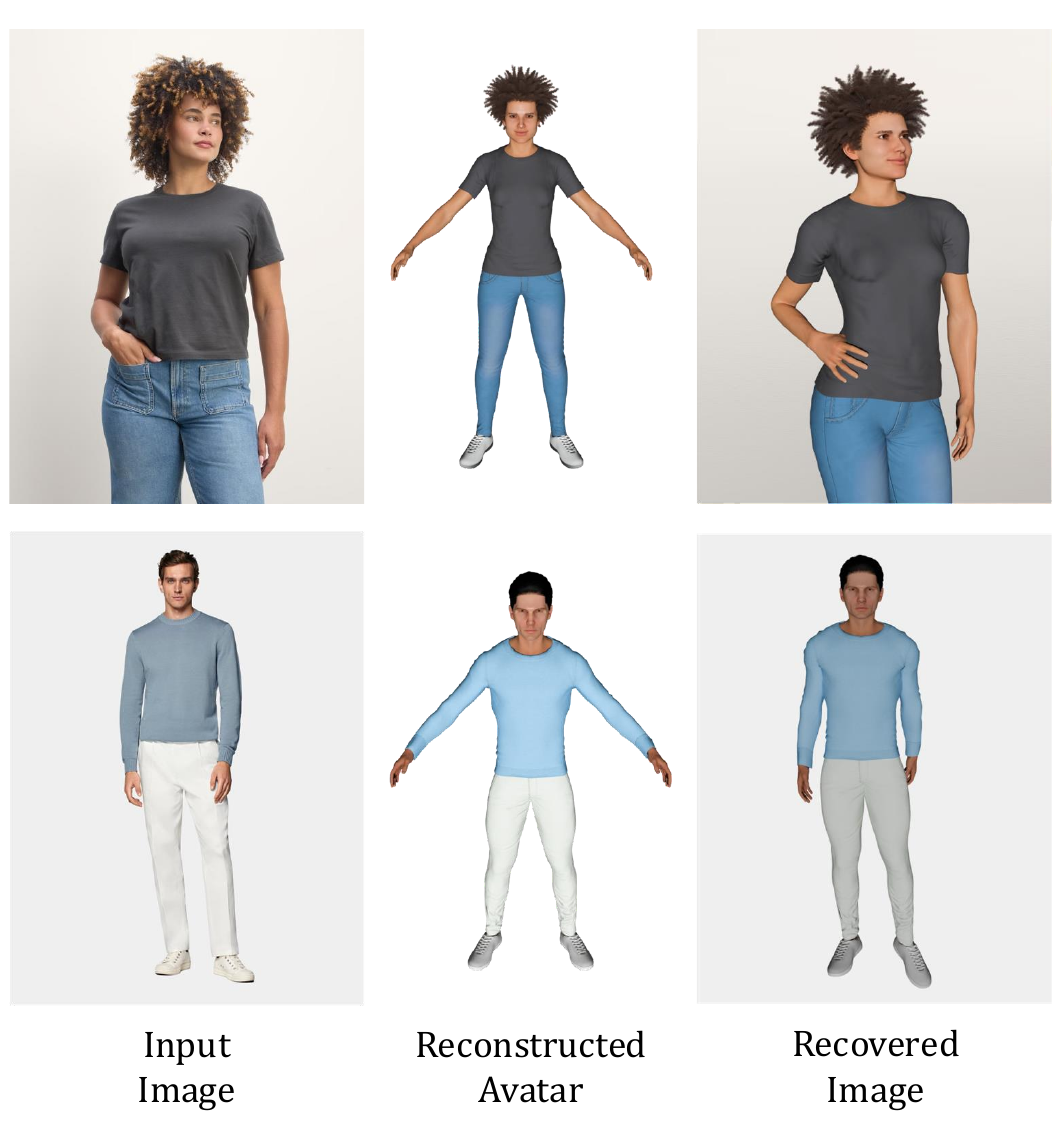}
    \vspace{-0.2in}
    \caption{Input images (left) and the output generated by AvatarForge (middle) showcasing reconstructed 3D avatars. The right images represent the recovered images. (On the other hand, manual effort is involved to achieve a similar visual effect.)}
    \vspace{-0.15in}
    \label{fig:image_input}
\end{figure}

\begin{figure}[t]
    \centering
    \includegraphics[width=\linewidth]{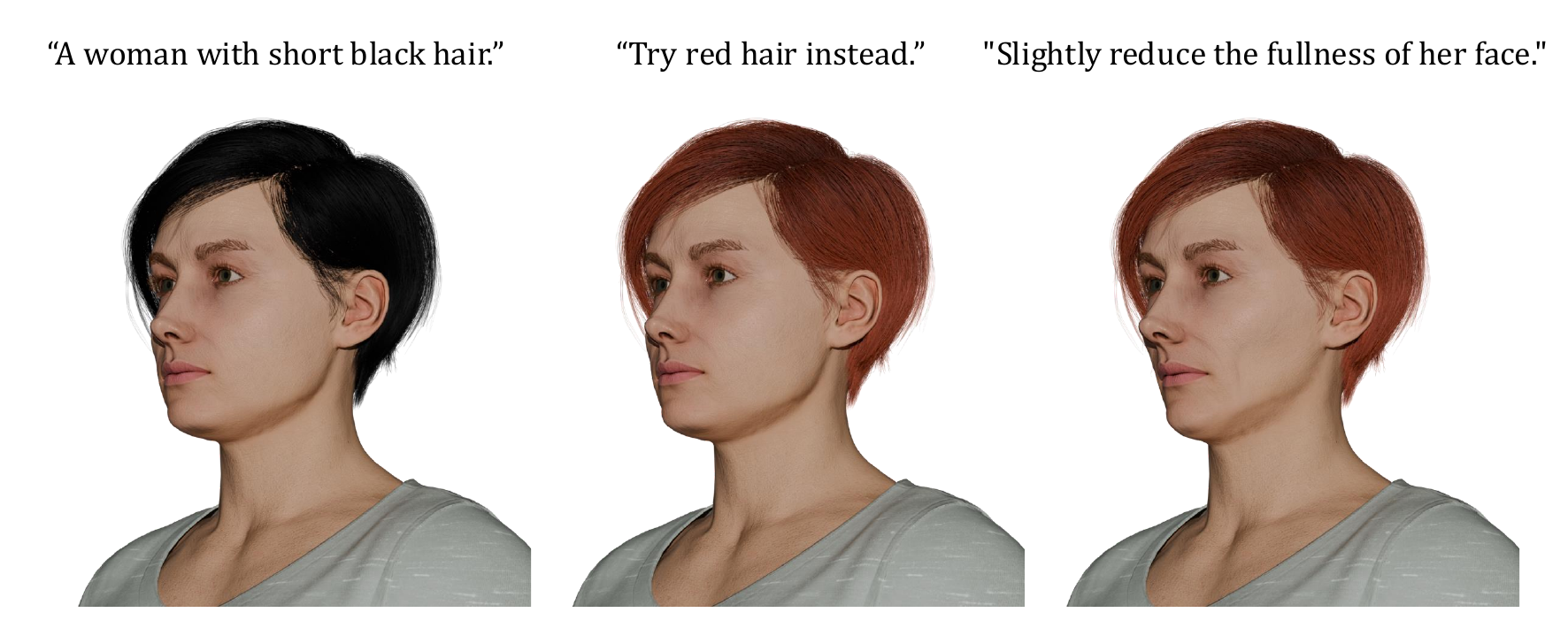}
    \vspace{-0.15in}
    \caption{Our approach allows users to edit the attributes of generated humans using natural language.}
    \vspace{-0.15in}
    \label{fig:editing}
\end{figure}

\begin{figure*}[t]
    \centering
    \includegraphics[width=0.9\linewidth]{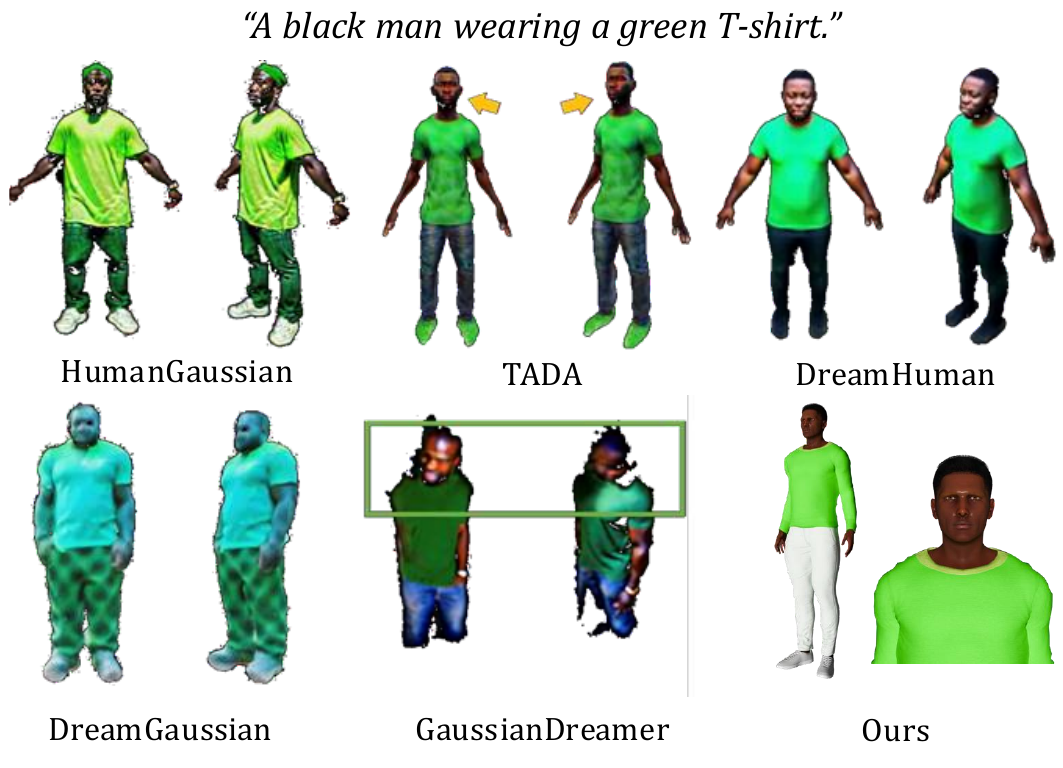}
    \vspace{-0.15in}
    \caption{Comparison of 3D avatar generation methods with state-of-the-art models. Our approach achieves improvements in avatar fidelity and produces more accurate and realistic avatars compared to the other models.}
    \vspace{-0.15in}
    \label{fig:comparison}
\end{figure*}

\section{Experimental Results}
In this section, we present a comprehensive evaluation of AvatarForge, showcasing its robust capabilities in generating highly customizable, realistic, and animatable 3D human avatars. We emphasize its versatility across various input modalities, including text and images, and demonstrate its clear advantages over current state-of-the-art methods.

\subsection{Diverse Avatar Generation}
Figure \ref{fig:gallery} illustrates the remarkable diversity of 3D human avatars produced by AvatarForge. The avatars span a wide range of body types, poses, and outfits, all generated based on detailed textual descriptions. This flexibility highlights AvatarForge's ability to adapt to diverse user specifications, enabling the creation of avatars that closely match real-world variety in human appearance. The resulting avatars not only display high realism but also incorporate intricate details, such as nuanced facial features and posture adjustments, underscoring the framework's capacity for fine-grained customization and high-quality representation.

\subsection{Image-to-Avatar Reconstruction}
In Figure \ref{fig:image_input}, we showcase the transformation of 2D input images into 3D avatars through AvatarForge. On the left, the input images depict simple 2D representations, which AvatarForge converts into detailed 3D models (middle). The right panel shows the avatars with minor manual refinements to enhance visual accuracy, emphasizing the system’s flexibility in refining generated content. This ability to reconstruct avatars from images is especially valuable for real-world applications such as digital twin creation, virtual character modeling, and other immersive experiences where 3D fidelity is crucial.

\subsection{Avatar Editing}
One of the key strengths of AvatarForge is its ability to not only generate 3D avatars but also to enable detailed editing of their attributes after reconstruction. Figure \ref{fig:editing} showcases the flexibility of the avatar editing process. After generating the initial 3D avatar from either text or image input, users can fine-tune the avatar’s appearance by specifying changes to attributes. The ability to iteratively modify avatars based on specific user requests allows for precise tailoring of the models to meet individual preferences, further demonstrating the strength of AvatarForge in the domain of customizable and interactive 3D avatar generation.

\subsection{Comparison with State-of-the-Art Models}
Figure \ref{fig:comparison} compares AvatarForge with existing leading models in the field of 3D human avatar generation. Our framework consistently outperforms these approaches by producing avatars that exhibit superior fidelity, capturing finer details in body structure, clothing, and poses. The comparison demonstrates AvatarForge's ability to generate more realistic and accurate representations, with particular strength in handling complex and varied textual descriptions and input sources. Unlike current models, which often struggle with the intricacies of human anatomy and customization, AvatarForge excels in delivering highly personalized, lifelike avatars with enhanced realism and expressiveness. This positions AvatarForge as a leading tool in the field, pushing the boundaries of what is possible in 3D human avatar creation.

\section{Limitation}
AvatarForge offers significant advantages over traditional deep 3D generation methods, including reduced computational load, improved surface geometry, and the ability to iteratively customize avatars through natural language commands. Its integration with off-the-shelf 3D generators enhances its versatility, making it adaptable for various creative needs. However, it has some limitations, such as less artistic detailing compared to specialized 3D generators and reliance on external tools for advanced features. While it excels at avatar creation and basic animation, its capabilities in high-end artistic rendering and complex motion control remain limited. Despite these challenges, AvatarForge provides a highly efficient and customizable solution for generating realistic and interactive avatars.

\section{Conclusion}
In this work, we introduced AvatarForge, a novel framework that enables the generation of highly customizable and animatable 3D human avatars from text and image inputs. By leveraging Large Language Models (LLMs) alongside advanced 3D human generation techniques, AvatarForge addresses key challenges in the creation of lifelike human avatars, particularly in terms of diversity, realism, and control over fine-grained body and facial details. Unlike existing methods that often lack precision or rely on limited datasets, AvatarForge offers a unique approach that combines user-driven customization with an auto-verification system, allowing for continuous refinement and high-fidelity avatar generation. Our results demonstrate that AvatarForge outperforms state-of-the-art methods in both text- and image-to-avatar generation, excelling in visual quality, customization flexibility, and ease of use. Moreover, the system’s ability to animate avatars via natural language commands significantly broadens its applicability across various domains, including gaming, film, and mixed reality environments. The major contribution of AvatarForge lies in its combination of AI-driven procedural generation with intuitive user interaction, providing a versatile tool that bridges the gap between artistic creation and technical modeling. Future work will focus on expanding AvatarForge's motion control capabilities and exploring its potential for new virtual and augmented reality applications.

{
    \small
    \bibliographystyle{ieeenat_fullname}
    \def\CVPR{IEEE/CVF Conference on Computer Vision and Pattern Recognition (CVPR)}\def\ECCV{ European Conference on Computer Vision (ECCV)}\def\ICCV{IEEE/CVF International Conference on Computer Vision (ICCV)}\def\NIPS{Advances in Neural Information Processing Systems (NeurIPS)}\def\ICML{International Conference on Machine Learning (ICML)}\def\ICLR{International Conference on Learning Representations (ICLR)}\def\WACV{IEEE/CVF Winter Conference on Applications of Computer Vision (WACV)}\def\CVPRW{IEEE/CVF Conference on Computer Vision and Pattern Recognition (CVPR) Workshops}\def\ICCVW{IEEE/CVF International Conference on Computer Vision (ICCV) Workshops}\def\ICRA{IEEE International Conference on Robotics and Automation (ICRA)}\def\TOG{ACM Transactions on Graphics (TOG)}\def\PAMI{IEEE Transactions on Pattern Analysis and Machine Intelligence (PAMI)}\def\TIP{IEEE Transactions on Image Processing (TIP)}\def\IJCV{International Journal of Computer Vision (IJCV)}\def\SIGGRAPH{ACM Transactions on Graphics
  (SIGGRAPH)}\def\SIGGRAPHASIA{ACM Transactions on Graphics (SIGGRAPH Asia)}\def\TOG{ACM Transactions on Graphics (TOG)}\def\threedv{International Conference on 3D Vision (3DV)}\def\TVCG{IEEE Transactions on Visualization and Computer Graphics (TVCG)}\def\PMLR{Proceedings of Machine Learning Research (PMLR)}

}

\end{document}